\documentclass[letterpaper, 10 pt, journal, twoside]{IEEEtran}

\IEEEoverridecommandlockouts                              

\usepackage{graphics} 
\usepackage[noadjust]{cite}
\usepackage{xcolor}
\usepackage{multirow}
\usepackage{caption}
\usepackage{graphicx}
\usepackage[hidelinks]{hyperref}
\usepackage[all]{hypcap}

\colorlet{red}{black}
\colorlet{magenta}{black}
\colorlet{green}{black}

\title{ReaDy-Go: Real-to-Sim Dynamic 3D Gaussian Splatting Simulation for Environment-Specific Visual Navigation with Moving Obstacles}

\author{Seungyeon Yoo$^{1}$, Youngseok Jang$^{2}$, Dabin Kim$^{1}$, Youngsoo Han$^{1}$, Seungwoo Jung$^{1}$, and H. Jin Kim$^{1}$
\thanks{Manuscript received: February, 14, 2026; Revised May, 7, 2026; Accepted June, 9, 2026.}
\thanks{This paper was recommended for publication by Editor P. Vasseur upon evaluation of the Associate Editor and Reviewers’ comments.
This work was supported by Samsung Research Funding \& Incubation Center of Samsung Electronics under Project Number SRFC-IT2402-17.} 
\thanks{$^{1}$Seungyeon Yoo, Dabin Kim, Youngsoo Han, Seungwoo Jung, and H. Jin Kim are with the Department of Aerospace Engineering, Seoul National University, Seoul, Republic of Korea
{\tt\footnotesize \{syeon.yoo, dabin404, cat7945, tmddn833, hjinkim\}@snu.ac.kr}}%
\thanks{$^{2}$Youngseok Jang is with the InnoCORE AI-Transformed Aerospace Research Center, KAIST, Daejeon, Republic of Korea
{\tt\footnotesize duscjs59@gmail.com}}%
\thanks{Digital Object Identifier (DOI): see top of this page.}
}

\begin{document}

\markboth{IEEE Robotics and Automation Letters. Preprint Version. Accepted June, 2026}
{YOO \MakeLowercase{\textit{et al.}}: ReaDy-Go} 

\maketitle

\begin{abstract}
Visual navigation models often struggle in real-world dynamic environments due to limited robustness to the sim-to-real gap and the difficulty of training policies tailored to target deployment environments (e.g., households, restaurants, and factories). Although real-to-sim navigation simulation using 3D Gaussian Splatting (GS) can mitigate these challenges, prior GS-based works have considered only static scenes or non-photorealistic human obstacles built from simulator assets, despite the importance of safe navigation in dynamic environments. To address these issues, we propose ReaDy-Go, a novel real-to-sim simulation pipeline that synthesizes photorealistic dynamic scenarios in target environments by augmenting a reconstructed static GS scene with dynamic human GS obstacles, and trains navigation policies using the generated datasets. The pipeline provides three key contributions: (1) a dynamic GS simulator that integrates static scene GS with a human animation module, enabling the insertion of animatable human GS avatars and the synthesis of plausible human motions from 2D trajectories, (2) a navigation dataset generation framework that leverages the simulator along with a robot expert planner designed for dynamic GS representations and a human planner, and (3) robust navigation policies to both the sim-to-real gap and moving obstacles. ReaDy-Go outperforms baselines across target environments in both simulation and real-world experiments, demonstrating improved navigation performance even after sim-to-real transfer and in the presence of moving obstacles. Moreover, zero-shot sim-to-real deployment in an unseen environment indicates its generalization potential. Project page: \href{https://syeon-yoo.github.io/ready-go-site/}{\texttt{https://syeon-yoo.github.io/ready-go-site/}}.
\end{abstract}


\section{Introduction}

\IEEEPARstart{V}{isual} navigation policies \textcolor{red}{that rely solely on} an RGB camera provide practical advantages \textcolor{red}{for robotic systems, including} reduced hardware complexity, lower sensor costs, and lighter payloads. Furthermore, \textcolor{red}{RGB-only navigation modules can be seamlessly integrated} with other vision-based perception and decision-making components. Despite these benefits, their performance in real-world dynamic environments often lacks robustness and efficiency, \textcolor{red}{limiting reliable deployment in practical settings.}

\textcolor{red}{The fundamental challenge lies in achieving robust navigation in environment-specific, dynamic real-world settings. RGB-only navigation models typically learn nonlinear visuomotor policies from high-dimensional monocular observations, where depth ambiguity complicates scene understanding. Therefore, most learning-based approaches are trained extensively in simulation, as collecting large-scale real-world navigation data is impractical. However, the resulting sim-to-real distribution gap significantly degrades performance during deployment.
Moreover, robots are commonly operated in environment-specific settings such as households, restaurants, or factories, where scene layouts are unique. General navigation models \cite{shah2022gnm, shah2023vint, 10610665} often fail to fully exploit environment-specific characteristics, which leads to reduced success rates \cite{Chhablani_2025_ICCV, NEURIPS2023_022ca1be}, while constructing digital twins for targeted dataset generation remains expensive \cite{ramakrishnan2021hm3d}.}

\begin{figure}[t]
  \centering
  \includegraphics[width=\linewidth]{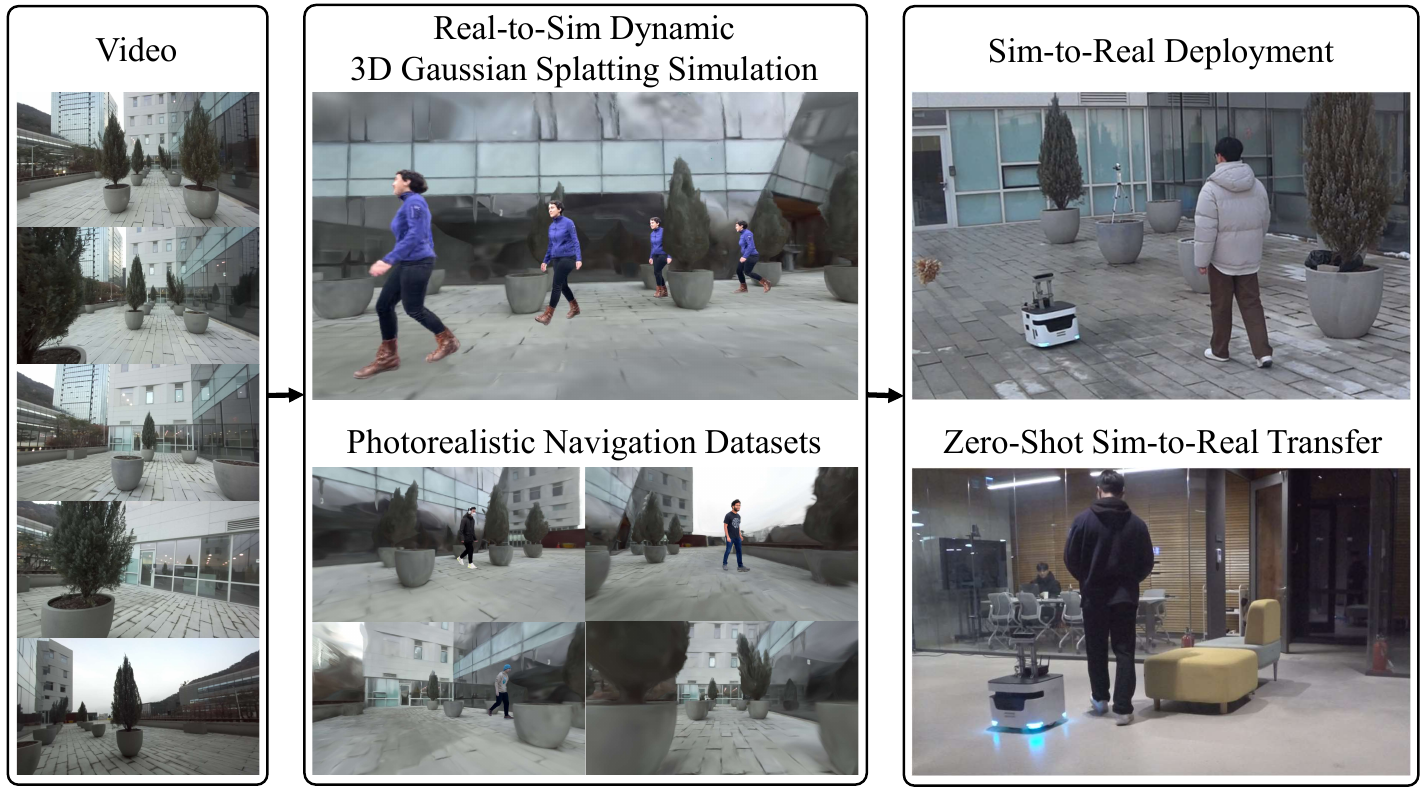}
  \caption{\textit{The proposed real-to-sim dynamic environment simulation pipeline for visual navigation.} ReaDy-Go generates photorealistic navigation datasets for dynamic scenarios and trains environment-specific visual navigation policies from these datasets. The resulting policies demonstrate robustness to the sim-to-real gap and moving obstacles.}
  \label{fig:thumbnail}
  \vspace{-0.7cm}
\end{figure}

To mitigate these limitations, recent works \cite{Chhablani_2025_ICCV, 11020756, quach2024gaussian, 10937041, miao2025performance, Xie_2025_CVPR} have proposed real-to-sim simulation pipelines based on 3D Gaussian Splatting (GS) \cite{kerbl3Dgaussians}. By reconstructing environments from RGB videos, GS enables high-fidelity rendering at fast frame rates, novel view synthesis, and simulation with an explicit 3D scene representation. GS-based pipelines leverage these advantages for navigation dataset generation and policy training to reduce the sim-to-real gap in target environments. However, existing GS-based navigation frameworks are limited to static scenes or introduce dynamic elements using non-photorealistic simulation assets, such as human meshes. Such limitations make it difficult to learn safe navigation in the presence of dynamic obstacles and to render photorealistic human appearances within reconstructed real-world environments. As a result, the generation of photorealistic navigation datasets for dynamic environments remains underexplored.

\textcolor{red}{Addressing this issue is essential for robot deployment in real-world environments with moving elements. In particular, there is currently no GS-based real-to-sim pipeline that jointly (1) models animatable human GS obstacles within reconstructed static GS scenes, (2) generates navigation datasets tailored to environment-specific training in dynamic settings, and (3) trains visual navigation policies that are robust under sim-to-real transfer and to moving obstacles.}

Motivated by these limitations, we propose ReaDy-Go, a photorealistic \textbf{Rea}l-to-Sim \textbf{Dy}namic 3D \textbf{G}aussian Splatting Simulati\textbf{o}n pipeline for environment-specific RGB-only visual navigation with moving obstacles (Fig.~\ref{fig:thumbnail}). The framework consists of three key components: (1) a dynamic GS simulator that integrates a static scene GS, an animatable human GS obstacle, and a human motion generation module, enabling the placement of a human in the scene and the synthesis of plausible motions conditioned on 2D trajectories; (2) a photorealistic dataset generation pipeline for dynamic environments that is composed of the dynamic simulator, a robot expert planner designed for dynamic GS representations, and a human planner; and (3) training an RGB-only visual navigation policy using imitation learning with the generated datasets.

To the best of our knowledge, ReaDy-Go is the first GS-based real-to-sim framework for visual navigation in dynamic environments that enables photorealistic dataset generation with human GS avatars and environment-specific navigation policy training. Our contributions are threefold.
\begin{itemize}
    \item Dynamic GS Simulator: We develop a photorealistic real-to-sim dynamic 3D Gaussian Splatting simulator with human GS obstacles. The simulator represents both the scene and the human as GS representations and generates human motion from a given 2D trajectory. \textcolor{magenta}{It unifies scene GS, animatable human GS avatars, and motion generation within a coherent dynamic simulator.}

    \item Photorealistic Dynamic Dataset Generation Pipeline: We propose a new pipeline, ReaDy-Go, that generates photorealistic dynamic scenario datasets and trains environment-specific visual navigation policies, without requiring mesh extraction and physics engine integration. \textcolor{magenta}{To enable dynamic navigation scenario generation, we develop a robot expert planner tailored to dynamic GS representations and a human planner, and formulate a unified pipeline that integrates these components with the dynamic GS simulator.}

    \item Robust Sim-to-Real Performance in Dynamic Environments: The proposed method achieves robust RGB-only visual navigation performance in dynamic environments in both simulation and sim-to-real transfer experiments. Furthermore, it shows generalization potential via zero-shot sim-to-real deployment in an unseen environment.
\end{itemize}

\section{Related Work}

\subsection{RGB-Only Visual Navigation Policies}
RGB-only navigation often uses learning-based methods to learn nonlinear visuomotor policies that map high-dimensional observations to actions while mitigating monocular depth ambiguity. Since learning-based navigation policies usually require large amounts of data, previous works have used simulation datasets to train policies by imitation learning or reinforcement learning \cite{9359345, 8877728, sadeghi2016cad2rl, NEURIPS2023_022ca1be, yadav2023ovrl}. To leverage the fact that simulation can easily capture multi-modal data, cross-modal learning that uses heterogeneous data during training while employing only RGB for inference has also been proposed to \textcolor{red}{distill multi-modal information into RGB-only policies} \cite{10542210, 9341049}. \textcolor{red}{Despite these advances, policies trained in simulation often suffer from sim-to-real performance degradation. Techniques such as domain randomization \cite{8877728, sadeghi2016cad2rl} may not fully eliminate distribution gaps.}

\textcolor{red}{An alternative direction} to address the sim-to-real gap for visual navigation policies is \textcolor{red}{to train} General Navigation Models (GNMs) on large-scale real-world navigation datasets \cite{10610665, shah2023vint, shah2022gnm}. GNMs show the potential to handle diverse robot embodiments and environments in a zero-shot manner using a single model. \textcolor{red}{However, they typically require a pre-built topological map (e.g., sequences of goal images) during inference and may underperform compared to environment-specific policies in target deployment settings \cite{Chhablani_2025_ICCV, NEURIPS2023_022ca1be}, as they cannot fully exploit scene-specific structural information.}

\textcolor{red}{To address these limitations, real-to-sim simulation approaches reconstruct target environments and generate photorealistic datasets tailored for environment-specific policy training. Scene reconstructions can be achieved using asset retrieval \cite{liu2025urbanverse, phone2proc} or 3D Gaussian Splatting, which provides explicit scene representations for rendering and planning.}

\subsection{GS-Based Real-to-Sim Simulation for Navigation Policies}
Real-to-sim simulation using 3D Gaussian Splatting (GS) has recently gained attention in robotics for reducing the sim-to-real gap in navigation \cite{Chhablani_2025_ICCV, 11020756, quach2024gaussian, 10937041, miao2025performance, Xie_2025_CVPR}. GS offers high-fidelity rendering at fast frame rates, novel view synthesis, and geometrically interpretable primitives useful for simulation.

The first type of work is rule-based planning methods using GS maps to exploit their geometric consistency and rendering quality \cite{10930696, 10870413}. While effective in static environments, these methods require access to the GS map at inference time and are not well suited for resource-constrained robotic platforms. In the second type of work, learning-based navigation policies are trained in GS-based real-to-sim environments and transferred to actual robot hardware such as drones \cite{quach2024gaussian, 10937041, miao2025performance}, wheeled robots \cite{Chhablani_2025_ICCV, Xie_2025_CVPR}, and legged robots \cite{11020756}. These approaches demonstrate improved sim-to-real transfer by leveraging photorealistic reconstructions of deployment environments.

Despite this advantage, existing GS-based real-to-sim navigation frameworks predominantly assume static scenes, whereas dynamic environments are essential for real-world applications. Moreover, some works simplify tasks by using predefined trajectories or strong goal cues (e.g., colored targets), reducing the task difficulty. Although Vid2Sim \cite{Xie_2025_CVPR} introduces dynamic obstacles via simulation assets, it may limit photorealism compared to fully GS-based pipelines and requires additional mesh extraction and physics engine integration to generate observations from heterogeneous representations (GS and meshes).

\begin{figure*}[t]
  \centering
  \includegraphics[width=\linewidth]{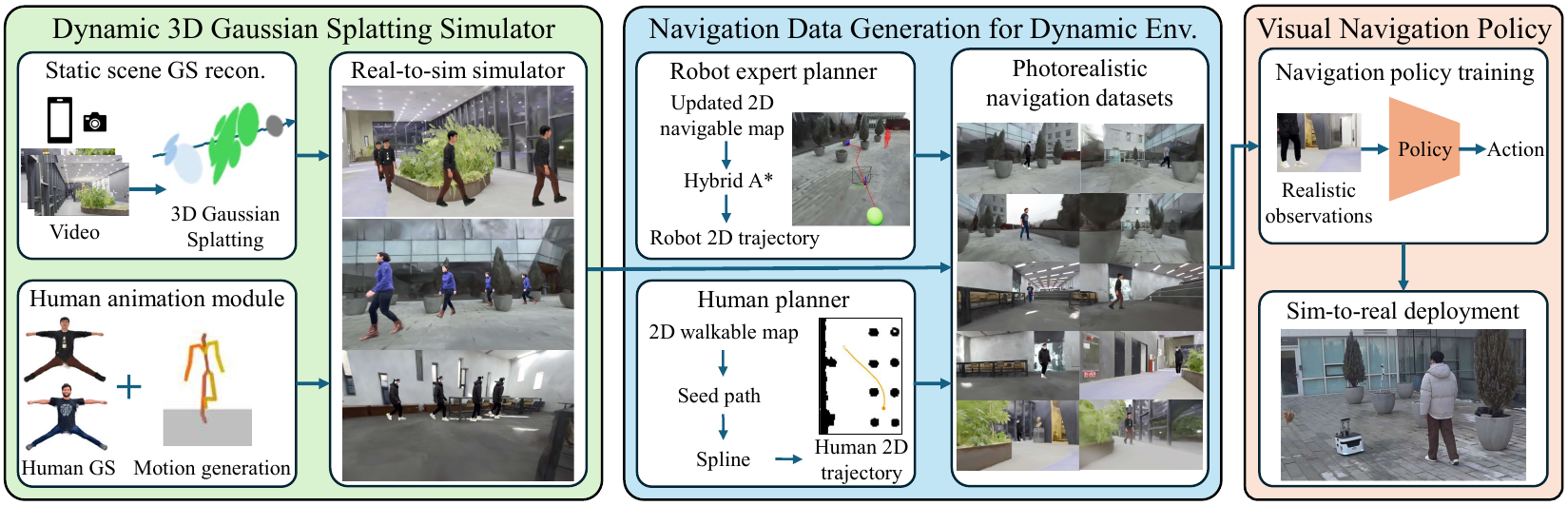}
  \vspace{-0.6cm}
  \caption{
  \textit{ReaDy-Go overview.} The proposed photorealistic simulation pipeline for visual navigation in dynamic environments consists of three main components: (1) a real-to-sim dynamic 3D Gaussian Splatting (GS) simulator with animatable human GS avatars, (2) photorealistic navigation dataset generation for dynamic scenarios, and (3) visual navigation policy training.
  }
  \label{fig:main}
  \vspace{-0.5cm}
\end{figure*}

In contrast, our method, ReaDy-Go, develops a pipeline for navigation policies that is composed of a photorealistic real-to-sim dynamic GS simulation using GS for both the scene and dynamic human obstacles, together with a robot expert planner and a human planner. We show that ReaDy-Go improves sim-to-real transfer performance of environment-specific navigation policies in dynamic environments and demonstrates policy generalization.

\section{Method}

Given a video of a static target deployment environment, ReaDy-Go generates photorealistic navigation datasets with moving human obstacles and trains an environment-specific navigation policy, as shown in Fig.~\ref{fig:main}. The pipeline consists of three main components: (1) a real-to-sim dynamic 3D Gaussian Splatting (GS) simulator, (2) dynamic navigation dataset generation using the simulator and planners, and (3) navigation policy training. \textcolor{magenta}{In particular, ReaDy-Go contributes a coherent formulation of a dynamic GS simulator that unifies static scene GS, animatable human GS avatars, and human motion generation. It also introduces a robot expert planner tailored to dynamic GS representations and a human planner, and integrates them with the dynamic GS simulator to form a unified data generation pipeline.} Each component is explained in detail in this section.

\subsection{A Photorealistic Real-to-Sim Dynamic GS Simulator}
\textcolor{magenta}{ReaDy-Go takes as input images from a monocular video and corresponding camera poses estimated at metric scale using COLMAP \cite{7780814} with an ArUco marker in the initial frames. It then builds a dynamic GS simulator by reconstructing a static GS scene and integrating it with a human animation module that places pre-extracted human GS models into the scene and animates them along desired 2D trajectories.}

\subsubsection{Static GS scene reconstruction}
The background scene in which the robot will be deployed is reconstructed using GS. GS is a representation that enables 3D geometry reconstruction, high-fidelity novel view synthesis, and fast training and rendering by fitting positions, rotations, scales, opacities, and colors of 3D Gaussian primitives to the training set images \cite{kerbl3Dgaussians}. Specifically, we employed PGSR \cite{10747190} for 3D scene reconstruction, which achieves high-quality surface reconstruction and rendering by compressing 3D Gaussians into flat planes and using geometric regularization loss terms in addition to a photometric loss. Improved geometric accuracy compared to the vanilla GS reduces Gaussian noise in the scene and improves multi-view consistency. This is important when extracting scene voxels and 2D occupancy grids for planners of the robot and dynamic obstacles.

\subsubsection{Human Animation Module}\label{sec:human_animation_module}
We set humans as dynamic obstacles in the real-to-sim GS scene. The human animation module places an animatable human GS model in the scene and then generates plausible human motion along a given obstacle trajectory.

Animatable human GS avatars are extracted from the NeuMan dataset \cite{jiang2022neuman} using HUGS \cite{kocabas2024hugs}. HUGS disentangles a dynamic human and a static scene from video and parameterizes the human GS in a canonical space initialized with the SMPL model \cite{SMPL:2015}. Leveraging these parameters, HUGS estimates human Gaussian attributes and linear blend skinning weights to animate the human GS under novel poses given SMPL joint parameters. The extracted human GS can be placed, animated, and rendered in novel GS scenes and viewpoints.

To generate natural human motion along desired 2D dynamic obstacle trajectories, PriorMDM \cite{shafir2024human} is adopted to predict the body root trajectory and joint angles in the SMPL parameters, which are used to animate human GS models. It enables fine-grained trajectory-level control over human motion using a motion diffusion model as a generative prior. Given a 2D trajectory, we convert it into body root linear and rotation velocities, normalize them to match the HumanML3D \cite{Guo_2022_CVPR} representation as the model input, and feed them into PriorMDM. Then, the predicted 3D body joint positions are fitted to SMPL parameters through SMPLify \cite{Bogo:ECCV:2016} and transformed into the world coordinate frame of the target environment.

\subsection{Navigation Data Generation for Dynamic Environments}
To generate photorealistic navigation datasets for scenarios in dynamic environments, ReaDy-Go proposes a pipeline that integrates our dynamic GS simulator, a robot expert planner designed for dynamic GS representations, and a human planner. By leveraging the simulator and planners, the pipeline collects RGB observations, actions, and relative goal positions as training samples for a navigation policy. This data generation process does not require onerous procedures such as scene mesh extraction and physics engine integration.

\subsubsection{Static scene voxelization with opacity filtering for planners}\label{sec:voxelization}
For planners, we first voxelize static scenes by marking a voxel as occupied if it contains all or part of the 1$\sigma$ Gaussian ellipsoid, following \cite{10930696}. However, reconstructed static GS scenes can contain spurious Gaussians around the ground, although they are suppressed by geometric regularization. Such noise hinders robot and human planning by reducing free space in the scene even though rendered images appear high-quality. For this reason, we filter out noise based on the accumulated opacities in each voxel during scene voxelization. If the sum of Gaussian opacities in a voxel does not exceed an opacity threshold, the voxel is classified as free space. This leads to more accurate free space regions around the ground, especially for weakly textured scenes. The filtered voxel map is converted to 2D occupancy maps for planners by projecting occupancy across height ranges: from near the ground to the robot height for the robot navigable map and from near the ground to the human height for the human walkable map. Occupied grids in each map are inflated by a safety margin, such as the robot or human radius, for safe planning.

\subsubsection{Robot expert planner for dynamic GS representations}
Collecting expert navigation data for policy training requires an expert planning algorithm capable of navigating dynamic GS environments while generating kinematically feasible trajectories for ground vehicles in our setting. However, prior planning algorithms designed for GS representations, such as Splat-Nav~\cite{10930696}, are restricted to static environments and rely on quadrotor-specific planning limited to holonomic vehicles. Similarly,  GaussNav~\cite{10870413} operates solely in static GS environments with a discrete action space, failing to utilize the full kinematic capabilities of ground vehicles. To overcome these limitations, we design an expert planner that can handle GS-based moving obstacles and generate feasible trajectories for ground vehicles by leveraging the Hybrid A$^{*}$ planner~\cite{dolgov2008practical} augmented with a motion primitive library.

\textcolor{magenta}{In the proposed planner, the robot navigable map is updated at each timestep to account for a dynamic obstacle if the human GS avatar appears within the camera field of view (FOV) and comes within a safety margin of the planned robot trajectory.} To obtain a cleaner set of human Gaussian primitives that better corresponds to the actual human-occupied region for 2D projection, we remove spurious primitives by discarding ellipsoids with high uncertainty, defined as those with a covariance trace exceeding the 75th percentile. We further downsample the remaining primitives based on their mean positions using voxel-grid downsampling. \textcolor{magenta}{Finally, these valid points are projected onto the 2D robot navigable map, where the human-occupied regions are then inflated using a safety buffer derived from a constant velocity prediction over a 2~s horizon to account for human movement.}

\begin{figure}[t]
  \centering
  \includegraphics[width=0.76\linewidth]{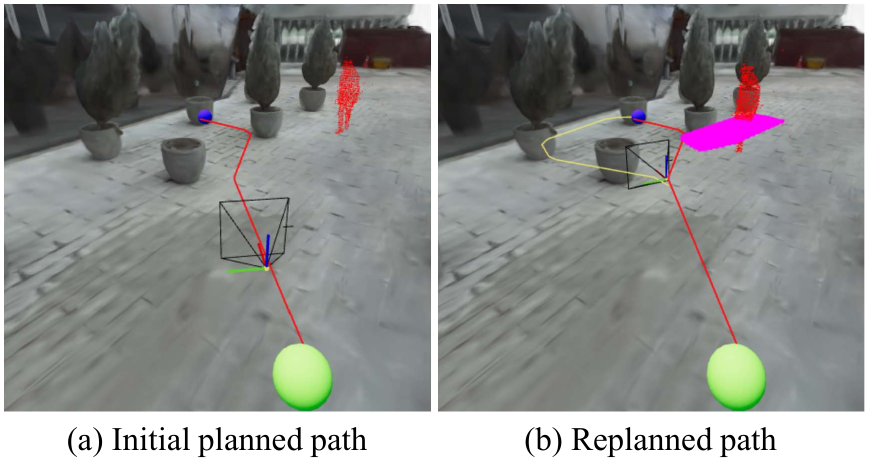}
  \vspace{-0.2cm}
  \caption{\textit{Visualization of the robot expert planner.} (a) The robot follows a collision-free path (red) from start (green) to goal (blue). (b) When a dynamic obstacle (human point cloud in red; inflated region in magenta) makes the path unsafe, the robot follows a replanned path (yellow).}
  \label{fig:expert}
  \vspace{-0.6cm}
\end{figure}

During the Hybrid A$^{*}$ expansion phase, each node transitions to its neighbors via a set of discrete motion primitives. These primitives are generated by combining three velocity scales (${1/3,2/3,1}$ of maximum velocity) with three steering commands (left, straight, right). The cost for each node is computed as a weighted sum of the primitive's arc length, steering penalty, and the heuristic cost-to-go to the goal. The expansion process terminates once a feasible path to the goal is identified. However, because the initial plan relies on short-horizon constant velocity predictions for human motion, the resulting trajectory may become unsafe during execution. To address this, we implement a reactive replanning mechanism. A replan is triggered either when a defined replanning period elapses or immediately if the minimum distance between a dynamic obstacle and the look-ahead segment of the executed trajectory falls below a safety margin. Fig.~\ref{fig:expert} illustrates the resulting path generated by the expert planner, highlighting the replanning behavior. With the proposed expert planner, we can generate safe robot trajectories in dynamic environments and gather expert demonstrations for navigation datasets.

\subsubsection{Human planner}
The human planner generates trajectories for dynamic obstacles using the 2D human walkable map. Firstly, to encourage human-robot interaction-rich scenarios, we sample pairs of human start and goal positions such that the resulting trajectories either cross or run parallel to the line connecting the robot's start and goal positions. A seed path is computed on the 2D map using a graph-based search algorithm, e.g., A$^\ast$. Then, a smooth trajectory is planned by spline-based optimization. Given this final 2D human trajectory, the human animation module generates SMPL motion parameters that enable a human to walk or run along the trajectory.

\subsubsection{Data generation}
With the planners, we can generate diverse dynamic scenarios in the proposed simulator. With a sufficient number of trials, the scenarios cover most regions in the scene and include various cases of static and dynamic obstacle avoidance. For each scenario, we record a set of tuples consisting of photorealistic RGB observations, corresponding expert outputs as actions (linear velocity $v$ and angular velocity $w$), and relative goal positions.

\subsection{Training Visual Navigation Policy}
Using generated photorealistic navigation datasets, we train end-to-end visual navigation policies using imitation learning to improve robustness to the sim-to-real gap and dynamic obstacles. To focus on validating the effect of ReaDy-Go simulation on navigation performance, we use a simple and lightweight architecture for the navigation policy, as described in Section~\ref{sec:architecture}. The RGB encoder receives three consecutive frames to incorporate ego history and the movement of a dynamic obstacle for safe and efficient obstacle detouring. Three consecutive latent vectors from the RGB encoder are concatenated with previous action ($v_{t-1}, w_{t-1}$) and relative goal positions in the robot local coordinate frame ($\Delta x_{t, r}, \Delta y_{t, r}$). The previous action provides ego movement information, and the relative goal positions are used for goal navigation. The concatenated vector passes through shallow fully connected layers and outputs the action ($v_t$, $w_t$), which is supervised to match the expert action using mean squared error.

\section{Experiments}
In this section, we evaluate the visual navigation performance in dynamic environments and the robustness to sim-to-real transfer of ReaDy-Go visual navigation policies trained in simulation. First, the qualitative results of ReaDy-Go simulation are examined to show the photorealistic real-to-sim dynamic GS simulation. Then, simulation and real-world experiments are conducted in static and dynamic tasks to compare its effective and robust navigation performance for target deployment environments against baselines. Finally, generalization of ReaDy-Go policies is investigated through navigation experiments in an unseen environment.

\subsection{Experimental Setup}
\subsubsection{Task description}
The task is point-goal navigation, where a robot should navigate from a start to a goal. For each episode, the start/goal pairs are sampled randomly in the free space of the environment. The robot should reach the goal without collisions within the scenario time limit. In the \textit{Static} task, the agent should avoid static obstacles in the scene. In the \textit{Dynamic} task, the agent should also detour around one or two dynamic obstacles in addition to the static obstacles. For each task and environment, we evaluate 100 episodes in simulation and 10 episodes in real-world experiments. \textcolor{green}{For each evaluation setting, the episode configurations are matched across the evaluated methods, including the same start/goal pairs and, in \textit{Dynamic} tasks, matched dynamic-obstacle trajectories.}

The robot used for real-world experiments was a differential wheeled robot equipped with a forward-facing fixed ZED2 camera. The robot was modeled as a unicycle and used an NVIDIA Jetson Orin NX for on-board inference. The camera obtained image observations for model input and wheel odometry was used to estimate relative goal positions.

\subsubsection{Evaluation metrics}
We evaluate navigation performance using Success Rate (SR) and Average Reaching Time (ART). SR is the proportion of successful scenarios among the total test scenarios, reflecting how the agent safely navigates in environments. A scenario is successful if the robot reaches the goal within 1 m before the maximum scenario length of 50 seconds. ART is the average time to finish scenarios across total scenarios, indicating how the agent effectively navigates to the goals. For failed scenarios, we set the reaching time to the maximum scenario length.

\subsubsection{Baselines}
We compare the following baselines against ReaDy-Go visual navigation policies to evaluate the effect of photorealistic dynamic GS simulation data for target deployment environments.
\begin{itemize}
    \item \textbf{Vid2Sim} \cite{Xie_2025_CVPR} generates real-to-sim navigation data by combining a GS scene, mesh, and a physics engine. Although it also generates dynamic environment scenarios, Vid2Sim uses a human simulation asset from the Unity engine as a dynamic obstacle, which is less photorealistic. \textcolor{magenta}{To isolate the effect of photorealistic dynamic obstacles on navigation policies, we employ the same policy architecture, human trajectories, and expert planner for both Vid2Sim and ReaDy-Go, while varying only the dynamic obstacle representation.}

    \item \textbf{GNM} \cite{shah2022gnm} is a general navigation model that is trained on large-scale navigation data. It can be deployed in diverse environments and across embodiments in a zero-shot manner. It uses image encoders and fully connected layers for goal-conditioned trajectory prediction.

    \item \textbf{ViNT} \cite{shah2023vint} extends GNM with a transformer architecture that improves cross-embodiment generalization and downstream adaptability.

    \item \textbf{NoMaD} \cite{10610665} extends ViNT by employing a diffusion model decoder to obtain a highly expressive policy for both goal-conditioned navigation and exploration.
\end{itemize}
ReaDy-Go and Vid2Sim are trained using datasets of dynamic scenarios and deployed in both \textit{Static} and \textit{Dynamic} tasks. For a fair comparison with image-goal navigation baselines (GNM, ViNT, and NoMaD), we provide them goal images captured at goal positions within 10 m of the start, with the camera oriented along the start-to-goal direction, which matches the robot's initial heading.

\begin{figure*}[t]
    \centering
    \includegraphics[width=0.72\textwidth]{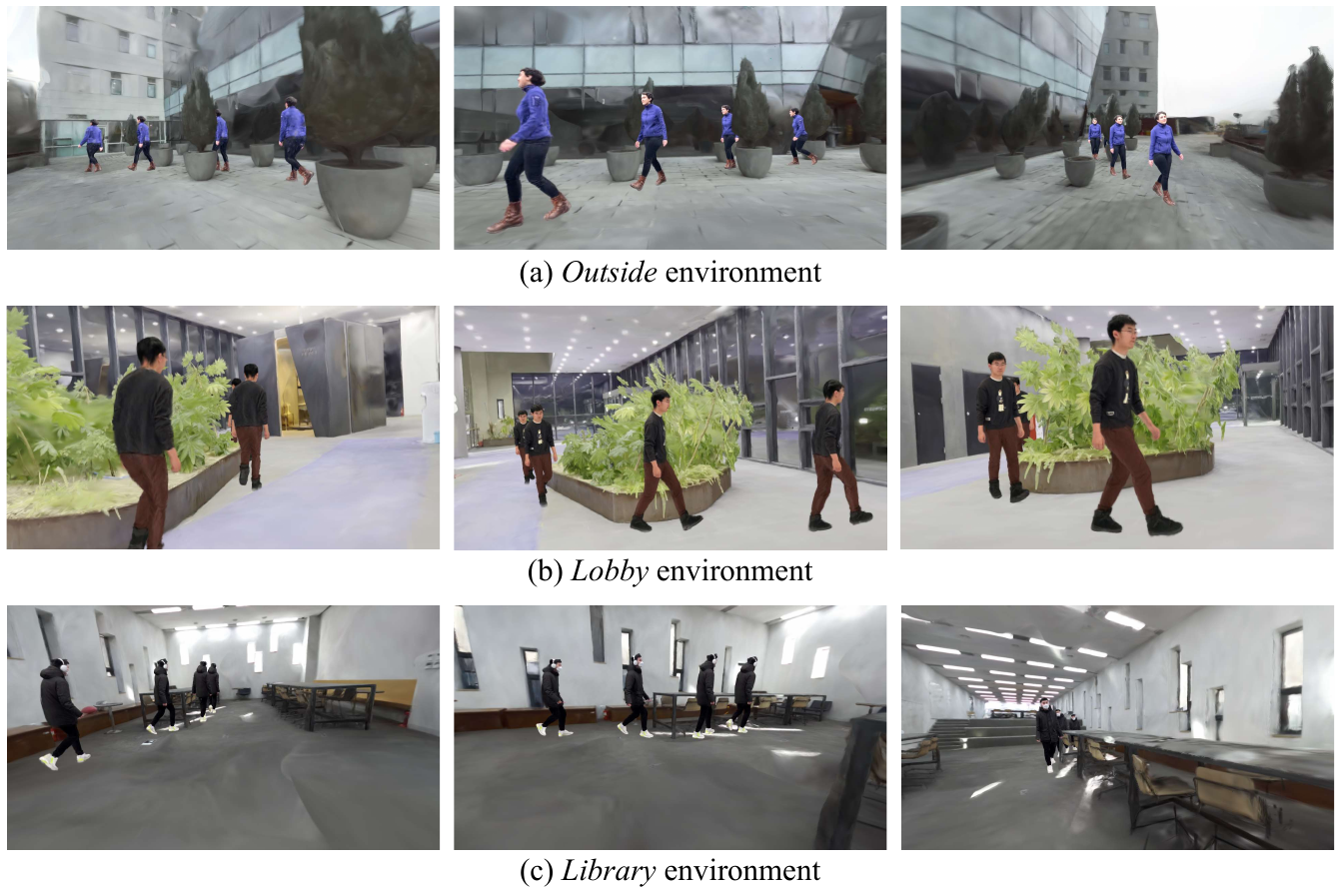}
    \vspace{-0.3cm}
    \caption{
    \textit{Qualitative novel-view synthesis results from the proposed dynamic GS simulation pipeline across diverse viewpoints and environments.} ReaDy-Go generates photorealistic, geometrically consistent dynamic scenarios with natural human motion from novel viewpoints, enabling navigation dataset generation for target deployment environments.
    }
    \label{fig:rendering}
    \vspace{-0.2cm}
\end{figure*}

\begin{table*}[t]
    \caption{Visual Navigation Performance in Simulation}
    \vspace{-0.3cm}
    \renewcommand{\arraystretch}{1.3}
    \begin{center}
    \resizebox{\linewidth}{!}{
    \begin{tabular}{ c | c | c  c  c  c | c  c  c  c | c  c  c  c }
    \hline
    \multirow{3}{*}{\centering Method} & \multirow{3}{*}{\centering Params} & \multicolumn{4}{|c|}{\textit{Outside}} & \multicolumn{4}{|c|}{\textit{Lobby}} & \multicolumn{4}{|c}{\textit{Library}} \\
    \cline{3-14}
     & & \multicolumn{2}{|c|}{\textit{Static}} & \multicolumn{2}{|c|}{\textit{Dynamic}} & \multicolumn{2}{|c|}{\textit{Static}} & \multicolumn{2}{|c|}{\textit{Dynamic}} & \multicolumn{2}{|c|}{\textit{Static}} & \multicolumn{2}{c}{\textit{Dynamic}} \\
     & & SR $\uparrow$ & \multicolumn{1}{c|}{ART (s) $\downarrow$} & SR $\uparrow$ & ART (s) $\downarrow$ & SR $\uparrow$ & \multicolumn{1}{c|}{ART (s) $\downarrow$} & SR $\uparrow$ & ART (s) $\downarrow$ & SR $\uparrow$ & \multicolumn{1}{c|}{ART (s) $\downarrow$} & SR $\uparrow$ & ART (s) $\downarrow$ \\
    \hline
    Expert & - & 100\% & \multicolumn{1}{c|}{-} & 91\% & - & 100\% & \multicolumn{1}{c|}{-} & 83\% & - & 100\% & \multicolumn{1}{c|}{-} & 78\% & - \\
    \hline
    GNM & 9M & 50\% & \multicolumn{1}{c|}{28.69} & 43\% & 31.89 & 50\% & \multicolumn{1}{c|}{28.12} & 50\% & 28.65 & 74\% & \multicolumn{1}{c|}{17.01} & 58\% & 23.48 \\
    NoMaD & 19M & 35\% & \multicolumn{1}{c|}{35.04} & 36\% & 34.64 & 25\% & \multicolumn{1}{c|}{38.89} & 21\% & 40.47 & 67\% & \multicolumn{1}{c|}{19.74} & 48\% & 27.94 \\
    ViNT & 30M & 50\% & \multicolumn{1}{c|}{28.72} & 36\% & 34.48 & 60\% & \multicolumn{1}{c|}{25.31} & 55\% & 27.82 & 75\% & \multicolumn{1}{c|}{15.85} & 62\% & 21.72 \\
    \hline
    Vid2Sim & 2M & 89\% & \multicolumn{1}{c|}{14.96} & 57\% & 26.88 & \textbf{98\%} & \multicolumn{1}{c|}{9.83} & 70\% & 21.06 & \textbf{91\%} & \multicolumn{1}{c|}{\textbf{9.82}} & 68\% & 19.92 \\
    \textbf{ReaDy-Go} & 2M & \textbf{90\%} & \multicolumn{1}{c|}{\textbf{13.43}} & \textbf{78\%} & \textbf{18.68} & \textbf{98\%} & \multicolumn{1}{c|}{\textbf{8.93}} & \textbf{78\%} & \textbf{17.59} & 86\% & \multicolumn{1}{c|}{11.08} & \textbf{80\%} & \textbf{13.98} \\
    \hline
    \end{tabular}
    }
    \end{center}
    \label{table_simulation}
    \vspace{-0.7cm}
\end{table*}

\subsection{Implementation Details}
\subsubsection{Dataset}
We selected three target environments, \textit{Outside}, \textit{Lobby}, and \textit{Library}, as shown in Fig.~\ref{fig:rendering}. Each environment was reconstructed as a GS scene using a monocular video recorded for about six minutes, \textcolor{red}{yielding} 1{,}000--1{,}500 images. \textcolor{red}{We used six human GS avatars (Section~\ref{sec:human_animation_module}) into the navigation datasets.} During ReaDy-Go simulation, we generated 400 training episodes for each environment, which corresponds to approximately 80k--120k data samples. The validation scenarios consist of 50 episodes for each environment, and we selected the checkpoint with the best validation performance, which is used for testing in simulation and the real world.

\subsubsection{Navigation policy training}\label{sec:architecture}
The navigation policy consists of ten convolutional layers with residual connections, which take an RGB input and output a 20-dimensional encoded vector, and three MLP layers with non-linear activations, which take a 64-dimensional concatenated vector (i.e., encoded vectors from three consecutive RGB images, the previous action, and the relative goal position). The policy predicts the action ($v$, $w$) and is trained with the Adam optimizer with a learning rate of $10^{-4}$. The image resolution is $144 \times 256$, and \textcolor{magenta}{the three consecutive images used by the policy are sampled at 0.5 s intervals. In real-world experiments, the policy requires only 18 ms per inference (55 Hz) on the on-board computer, while the camera operates at 20 Hz.}

\subsection{Visualization of ReaDy-Go Dynamic GS Simulation}
\textcolor{red}{Qualitative results demonstrate that ReaDy-Go produces photorealistic real-to-sim dynamic environments, as shown in Fig.~\ref{fig:rendering}. First, the proposed human animation module generates plausible body motions for human GS avatars within static GS scenes along given 2D trajectories, without relying on a physics engine. Second, the framework supports scalable generation of thousands of photorealistic navigation scenarios via expert and human planners. Finally, novel view synthesis results confirm that geometric consistency can be preserved from arbitrary viewpoints in dynamic settings.}

\begin{table*}[t]
    \caption{Real-World Visual Navigation Performance}
    \vspace{-0.3cm}
    \renewcommand{\arraystretch}{1.3}
    \begin{center}
    \resizebox{\linewidth}{!}{
    \begin{tabular}{ c | c | c  c  c  c | c  c  c  c | c  c  c  c }
    \hline
    \multirow{3}{*}{\centering Method} & \multirow{3}{*}{\centering Params} & \multicolumn{4}{|c|}{\textit{Outside}} & \multicolumn{4}{|c|}{\textit{Lobby}} & \multicolumn{4}{|c}{\textit{Library}} \\
    \cline{3-14}
     & & \multicolumn{2}{|c|}{\textit{Static}} & \multicolumn{2}{|c|}{\textit{Dynamic}} & \multicolumn{2}{|c|}{\textit{Static}} & \multicolumn{2}{|c|}{\textit{Dynamic}} & \multicolumn{2}{|c|}{\textit{Static}} & \multicolumn{2}{c}{\textit{Dynamic}} \\
     & & SR $\uparrow$ & \multicolumn{1}{c|}{ART (s) $\downarrow$} & SR $\uparrow$ & ART (s) $\downarrow$ & SR $\uparrow$ & \multicolumn{1}{c|}{ART (s) $\downarrow$} & SR $\uparrow$ & ART (s) $\downarrow$ & SR $\uparrow$ & \multicolumn{1}{c|}{ART (s) $\downarrow$} & SR $\uparrow$ & ART (s) $\downarrow$ \\
    \hline
    ViNT & 30M & 50\% & \multicolumn{1}{c|}{33.54} & 30\% & 39.22 & 60\% & \multicolumn{1}{c|}{27.83} & 20\% & 42.34 & 80\% & \multicolumn{1}{c|}{20.60} & 40\% & 34.96 \\
    Vid2Sim & 2M & 90\% & \multicolumn{1}{c|}{17.72} & 60\% & 28.98 & 70\% & \multicolumn{1}{c|}{25.35} & 40\% & 36.50 & 90\% & \multicolumn{1}{c|}{15.44} & 60\% & 28.54 \\
    \textbf{ReaDy-Go} & 2M & \textbf{100\%} & \multicolumn{1}{c|}{\textbf{17.45}} & \textbf{90\%} & \textbf{20.68} & \textbf{90\%} & \multicolumn{1}{c|}{\textbf{18.93}} & \textbf{70\%} & \textbf{26.49} & \textbf{100\%} & \multicolumn{1}{c|}{\textbf{11.12}} & \textbf{80\%} & \textbf{19.57} \\
    \hline
    \end{tabular}
    }
    \end{center}
    \label{table_realworld}
    \vspace{-0.5cm}
\end{table*}

\begin{figure*}[t]
  \centering
  \includegraphics[width=0.92\textwidth]{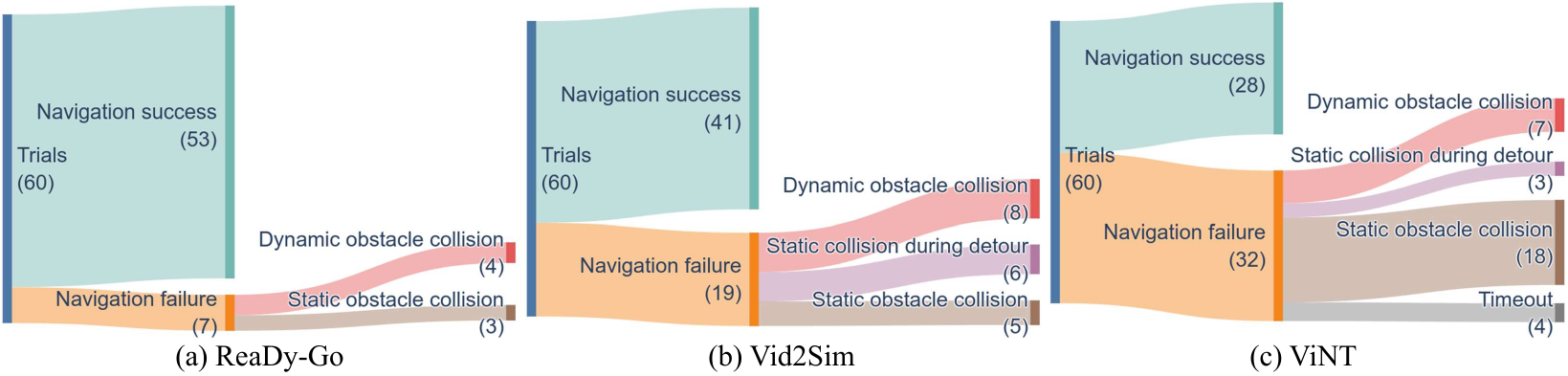}
  \vspace{-0.2cm}
  \caption{\textcolor{magenta}{\textit{Failure case analysis in real-world experiments.} ReaDy-Go yields fewer failures than the baselines, especially in failure modes related to dynamic obstacle avoidance, including \textit{Dynamic obstacle collision} and \textit{Static collision during detour}.}}
  \label{fig:failures}
  \vspace{-0.5cm}
\end{figure*}

\subsection{Visual Navigation Performance in Simulation}
The impact of photorealistic dynamic GS simulation datasets on visual navigation policies is examined through simulation tests comparing ReaDy-Go against baselines, as summarized in Table~\ref{table_simulation}. We observe two key takeaways from the experiments. First, training environment-specific navigation policies using real-to-sim simulation is crucial for safe and efficient navigation. ReaDy-Go and Vid2Sim, both trained in real-to-sim target environments, achieve higher success rates and lower average reaching times than general navigation models (GNM, NoMaD, and ViNT) across all tasks and environments. This performance is particularly notable given that the general models require up to $15\times$ more parameters. These indicate that real-to-sim simulation with GS is a cost-effective and scalable approach to achieve fewer collisions and faster task completion with only a video.

Second, photorealistic dynamic obstacles in our simulation are a key factor in maintaining visual navigation performance in dynamic environments. While ReaDy-Go and Vid2Sim perform comparably in static environments, their performance differs significantly in dynamic settings. ReaDy-Go maintains high success rates and low average reaching times relatively well, whereas Vid2Sim exhibits a substantial performance drop. Since the two methods differ only in the training data, i.e., photorealistic human GS dynamic obstacles for ReaDy-Go versus human assets in a physics engine for Vid2Sim, these results suggest that the proposed photorealistic dynamic GS simulation helps improve navigation performance in dynamic scenarios, which is a practical advantage for robot deployment.

\begin{figure}[t]
  \centering
  \includegraphics[width=\linewidth]{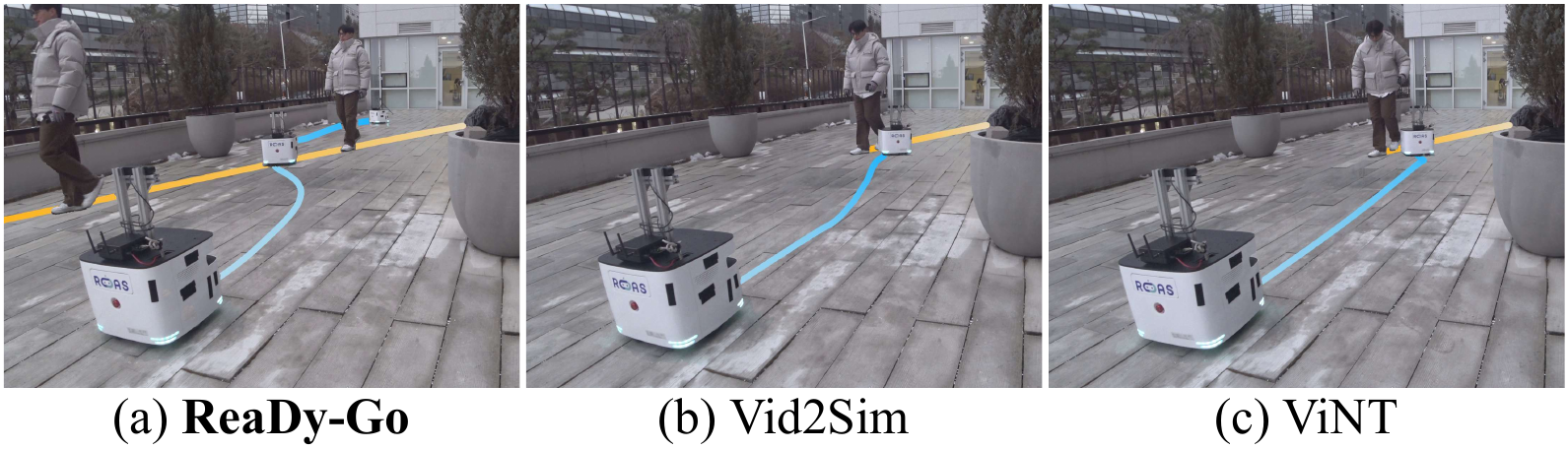}
  \vspace{-0.6cm}
  \caption{\textcolor{green}{\textit{Qualitative real-world navigation results.} ReaDy-Go avoids a dynamic obstacle, while the baselines collide with it.}}
  \label{fig:qualitative_real_world}
  \vspace{-0.5cm}
\end{figure}

\subsection{Real-World Visual Navigation Performance}
To investigate the effectiveness of the proposed pipeline in mitigating the sim-to-real gap, we compare the real-world navigation performance of ReaDy-Go and baselines in the target environments \textcolor{magenta}{with six previously unseen humans.} For baselines in real-world experiments, Vid2Sim and ViNT are used, as they achieve better performance among the four baseline models in the simulation experiments.

Table~\ref{table_realworld} and \textcolor{green}{Fig.~\ref{fig:qualitative_real_world}} report three main findings. First, the photorealistic real-to-sim simulation of ReaDy-Go facilitates sim-to-real transfer of visual navigation policies trained solely in simulation. ReaDy-Go achieves comparable success rates in both \textit{Static} and \textit{Dynamic} in the real world, consistent with its simulation results across all environments, even though the real-world environments exhibit minor appearance changes between data collection and deployment (e.g., floor texture, lighting, or background variations). This indicates that photorealistic datasets for dynamic scenarios improve the robustness of end-to-end visual navigation policies to the sim-to-real gap.

Second, environment-specific policies (ReaDy-Go and Vid2Sim) trained on real-to-sim datasets for target environments achieve higher success rates and lower average reaching times than the general navigation model (ViNT), although ViNT requires $15\times$ more parameters. This validates the practicality of ReaDy-Go for robots operating in specific environments, such as households, restaurants, and factories.

Lastly, photorealistic dynamic obstacles in ReaDy-Go are important for maintaining navigation performance in dynamic environments. \textcolor{green}{As in simulation, ReaDy-Go and Vid2Sim achieve comparable success rates in \textit{Static}, but their performance diverges in \textit{Dynamic}.} ReaDy-Go shows the highest success rate and the lowest average reaching time in \textit{Dynamic}, with only a slight performance degradation compared to \textit{Static}. In contrast, Vid2Sim exhibits a larger performance drop in \textit{Dynamic} than ReaDy-Go. \textcolor{magenta}{Because ReaDy-Go and Vid2Sim differ only in the representation of dynamic obstacles, this result suggests that photorealistic human GS avatars improve navigation robustness in dynamic environments compared to the non-photorealistic human meshes used in Vid2Sim.}

\textcolor{magenta}{An analysis of the failure cases in Fig.~\ref{fig:failures} confirms these findings. First, ViNT exhibited substantially more collisions with static obstacles than ReaDy-Go and Vid2Sim. This observation suggests that training an environment-specific visual navigation policy provides a practical advantage for real-world deployment. Second, while ReaDy-Go and Vid2Sim showed similar numbers of failures in cases unrelated to dynamic obstacle interactions, ReaDy-Go was more robust in situations involving dynamic obstacles. These results highlight the importance of incorporating photorealistic dynamic obstacles into the simulation pipeline to achieve robust navigation performance in dynamic real-world environments.}

\begin{table}[t]
    \caption{ReaDy-Go Generalization to an Unseen Env.}
    \vspace{-0.3cm}
    \renewcommand{\arraystretch}{1.3}
    \begin{center}
    \begin{tabular}{ c | c c }
    \hline
    Task & SR $\uparrow$ & ART (s) $\downarrow$ \\
    \hline
    \textit{Static} & 70\% & 30.60 \\
    \textit{Dynamic} & 50\% & 35.45 \\
    \hline
    \end{tabular}
    \end{center}
    \label{table_generalization}
    \vspace{-0.6cm}
\end{table}

\subsection{Zero-Shot Sim-to-Real Generalization}
We further \textcolor{red}{assess} the generalization potential of ReaDy-Go by measuring zero-shot sim-to-real performance in an unseen environment. For this experiment, the policy is trained on the combined datasets from three environments, i.e., a total of 1,200 episodes from \textit{Outside}, \textit{Lobby}, and \textit{Library}. We then deploy the policy to an unseen environment, as illustrated in Fig.~\ref{fig:thumbnail}. The policy achieves over a 50\% success rate in both \textit{Static} and \textit{Dynamic}, with a higher average reaching time than in the training environments, as depicted in Table~\ref{table_generalization}. These results \textcolor{red}{suggest} that the policy learns general navigation behaviors, such as detouring around static and dynamic obstacles while reaching the goal, but requires a longer time to complete the task, as the policy does not have prior training in the unseen scene. ReaDy-Go \textcolor{red}{can provide} a scalable approach toward general navigation if trained on \textcolor{red}{larger} navigation datasets from diverse environments, which can be easily reconstructed from a few minutes of videos using the proposed pipeline.

\section{Conclusion}
In this work, we propose ReaDy-Go, a real-to-sim dynamic GS simulation pipeline for training visual navigation policies. By integrating a GS scene, a human animation module, an expert planner, and a human planner, it generates photorealistic navigation datasets for dynamic scenarios in target environments. 
The navigation policy trained on the datasets achieves better navigation performance in target environments compared to baseline methods in both simulation and the real world.
The results confirm that ReaDy-Go facilitates robust sim-to-real transfer even in dynamic environments and highlights the advantages of environment-specific policies and photorealistic dynamic obstacles. Furthermore, ReaDy-Go shows generalization potential through zero-shot sim-to-real deployment in an unseen environment. \textcolor{magenta}{Future work includes expanding training environments to strengthen generalization and integrating safe reinforcement learning methods to handle more challenging scenarios involving diverse dynamic agents beyond humans, dense dynamic obstacle settings, and aggressive human motions.}

\bibliographystyle{IEEEtranBST/IEEEtran}
\bibliography{reference, IEEEtranBST/IEEEabrv}

\end{document}